\documentclass[runningheads]{llncs}
\usepackage[T1]{fontenc}
\usepackage{graphicx}
\usepackage{booktabs}
\usepackage[misc]{ifsym}
\newcommand{\corr}{(\Letter)}
\usepackage{mwe}
\usepackage{mwe}
\usepackage{amsmath}
\usepackage{amsfonts}
\usepackage{subfigure}

\usepackage[usenames,dvipsnames]{xcolor}
\usepackage{tcolorbox}
\usepackage{tabularx}
\usepackage{array}
\usepackage{colortbl}
\tcbuselibrary{skins}

\newcolumntype{Y}{>{\raggedleft\arraybackslash}X}

\tcbset{tab1/.style={fonttitle=\bfseries\large,fontupper=\normalsize\sffamily,
colback=yellow!10!white,colframe=blue!75!black,colbacktitle=Salmon!40!white,
coltitle=black,center title,freelance,frame code={
\foreach \n in {north east,north west,south east,south west}
{\path [fill=blue!75!black] (interior.\n) circle (3mm); };},}}

\tcbset{tab2/.style={enhanced,fonttitle=\bfseries,fontupper=\normalsize\sffamily,
colback=yellow!10!white,colframe=blue!50!black,colbacktitle=Salmon!40!white,
coltitle=black,center title}}

\usepackage{amsmath,amssymb}
\usepackage{bbm}

\usepackage{booktabs}

\usepackage{subcaption}
\usepackage{float}

\usepackage{hyperref,cleveref}

\begin{document}

\title{Diving Deep: Forecasting Sea Surface Temperatures and Anomalies}


\author{Ding Ning\inst{1} \and
Varvara Vetrova\inst{1} \corr \and
Karin R. Bryan\inst{2} \and
Yun Sing Koh\inst{2} \and
Andreas Voskou\inst{3} \and
N'Dah Jean Kouagou\inst{4} \and
Arnab Sharma\inst{4}
}

\authorrunning{D. Ning et al.}

\institute{University of Canterbury, New Zealand \email{ding.ning@pg.canterbury.ac.nz,varvara.vetrova@canterbury.ac.nz}
\and
University of Auckland, New Zealand \email{\{karin.bryan,y.koh\}@auckland.ac.nz}
\and
Cyprus University of Technology, Cyprus \email{ai.voskou@edu.cut.ac.cy}
\and
Paderborn University, Germany \email{\{ndah.jean.kouagou,arnab.sharma\}@upb.de}
}

\maketitle              

\begin{abstract}
This overview paper details the findings from the Diving Deep: Forecasting Sea Surface Temperatures and Anomalies Challenge at the European Conference on Machine Learning and Principles and Practice of Knowledge Discovery in Databases (ECML PKDD) 2024. The challenge focused on the data-driven predictability of global sea surface temperatures (SSTs), a key factor in climate forecasting, ecosystem management, fisheries management, and climate change monitoring. The challenge involved forecasting SST anomalies (SSTAs) three months in advance using historical data and included a special task of predicting SSTAs nine months ahead for the Baltic Sea. Participants utilized various machine learning approaches to tackle the task, leveraging data from ERA5. This paper discusses the methodologies employed, the results obtained, and the lessons learned, offering insights into the future of climate-related predictive modeling.

\end{abstract}

\section{Problem Definition}

Hurricanes, mass coral bleaching, disruption of sea mammal migration patterns, and extreme weather events, such as unusually hot summers or cold winters, all share a common driver: temperature changes in our seas and oceans. The goal of this challenge is to investigate the predictability of global sea surface temperatures (SSTs). Predicting SSTs is crucial for several reasons:
\begin{itemize}
\item \textbf{Climate forecasting}. SSTs are key indicators of climate patterns and help predict weather phenomena such as hurricanes, droughts, and floods. Understanding SST variations enables scientists to anticipate and prepare for extreme weather events more effectively.
\item \textbf{Ecosystem management}. SSTs influence marine ecosystems, affecting species distribution, migration patterns, and reproductive cycles. Accurate SST predictions assist marine biologists and conservationists in effectively managing and protecting marine biodiversity.
\item \textbf{Fisheries management}. Many fish species depend on specific temperature ranges for breeding, feeding, and migration. Predicting SSTs allows fisheries managers to make informed decisions about quotas, fishing seasons, and habitat conservation, ensuring sustainable fish populations.
\item \textbf{Human activities}. SSTs impact various human activities, including shipping, tourism, and coastal development. Accurate SST predictions enable stakeholders to plan and adapt infrastructure, coastal defenses, and tourism activities in response to changing ocean temperatures.
\item \textbf{Climate change monitoring}. SSTs are critical indicators of climate change, with rising temperatures affecting ocean circulation, sea levels, and weather patterns. Accurate SST predictions help scientists monitor and assess the impacts of climate change on the oceans and the broader environment.
\end{itemize}

Variability in SSTs, also known as SST anomalies (SSTAs), is linked to climate oscillations and occurrences of extreme events, including the El Niño‐Southern Oscillation (ENSO), the Indian Ocean Dipole (IOD) oscillation, and marine heatwaves. In this challenge, the participants specifically focus on forecasting SSTAs for three months in advance.

Here, we present highlights and lessons learned in the Diving Deep Challenge at the European Conference on Machine Learning and Principles and Practice of Knowledge Discovery in Databases (ECML PKDD) 2024.

\section{Challenge Setup}

The challenge is structured according to the following components.

\subsection{Data and Task}

The SSTA data, along with additional features such as sea surface temperatures (SSTs), mean sea level pressures (MSLPs), and air temperatures at two meters above the surface (T2Ms), are sourced from ERA5, the fifth-generation reanalysis conducted by the European Centre for Medium‐Range Weather Forecasts (ECMWF) \cite{hersbach2020era5}. This dataset covers the past nine decades globally and provides monthly estimates of various atmospheric, land, and oceanic variables with a spatial resolution of 0.25°, spanning from January 1940 to the present.

The SSTA dataset is derived by subtracting a climatology value for each month from the corresponding SST values. Here, climatology is defined as the average SST over a specified period. Each column in the provided CSV files contains a time series of SSTAs, SSTs, MSLPs, or T2Ms for a specific location, with the coordinates for each location also included in a separate coordinate CSV file. The output CSV contains SSTA values shifted three months in advance for the same locations.

The task is to predict SSTAs three months in advance using previous SSTA, SST, MSLP, and T2M values. The data are split into a training set and a test set based on the temporal domain. For the training set, SSTA, SST, MSLP, and T2M data at time steps from January 1940 to December 2010 are provided. Participants are encouraged to define the inputs and outputs independently based on their chosen methods. In the test set, inputs only for time steps from January 2011 to September 2023 are supplied, with an initial window size of 12. The outputs from April 2011 to December 2023 are hidden and used to compute the evaluation score. The 142 test input and output combinations are shuffled and divided into two subdatasets, each with 71 combinations.

Additionally, we ask participants to make a more challenging prediction for the SSTAs at three locations in the Baltic Sea for September 2024, with the coordinates: (19.6875°E, 56°N), (19.6875°E, 58°N), and (19.6875°E, 62°N). Using December 2023 as the last input time step, this additional challenge requires a nine-month-ahead forecast, with the answer not revealed until October 2024.
We are interested in observing how participants approach longer-term forecasting and exploring the practical use of SSTA forecasts.

\subsection{Evaluation}

The evaluation is based on the provided SSTA data, structured as follows: each column in the CSV file contains blocks of 12 time series of SSTA. The objective is to produce a three-month-ahead SSTA forecast for each block. For example, if the first block contains a time series of January 2011, February 2011, …, December 2011, the forecast should be for March 2012. Note that the dataset omits timestamps.

The evaluation metric is calculated as the difference between the RMSE of a simple baseline and the RMSE of the participant's forecast, averaged across all locations. The baseline model is the persistence model, where the current SSTA value is used as the three-month-ahead forecast.  Winners were selected based on their solution, the reports provided, and their prediction in the Baltic Sea location for September 2024.

Additional support materials can be found in \cite{ning2024harnessing}.

\section{Solutions}

This section includes the solutions of the top two winners.

\subsection{Solution by Team randomguy}
\subsubsection{Preliminary analysis}
\begin{figure}[htbp]
    \centering
    \subfigure{\includegraphics[scale=0.403]{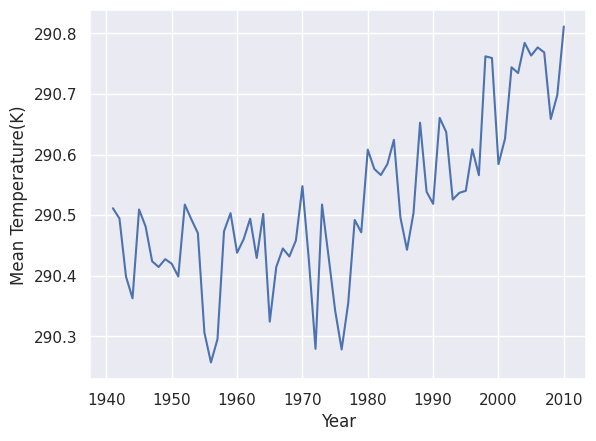}}
    \subfigure{\includegraphics[scale=0.403]{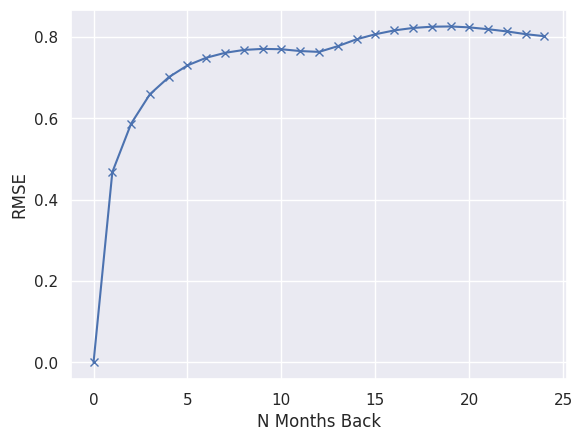}}
    \caption{\textbf{Left}: Global average temperature per year, \textbf{Right}: RMSE using N months back values as estimation}
    \label{fig:side_by_side}
\end{figure}

The first step was analyzing the data to gain a deeper understanding of the task, the goal, and the available data. On the left side of Figure \ref{fig:side_by_side}, we present the global average temperature per year. From this figure, we observe that the training data start to change rapidly after the 1980s-1990s, with a consistently increasing sea temperature. On the right side, we present the performance of a persistence model predicting N months ahead. As is apparent, performance drops rapidly for small
N values and for $N>6$ reaches a slower, periodic/seasonal phase.

In our preliminary modeling attempts, we first employed global deep learning models, experimenting with several MLP and 2D convolutional networks. These models failed to outperform the persistent model independently. However, when used to correct the persistent model, they showed slight improvement. Although more advanced techniques like graph neural networks (GNN) \cite{ning2024harnessing} exist, time constraints prevented their proper exploration.

In a different approach, instead of using a global model, we treated each target point as a separate row, converting the task into a tabular data problem and employing relevant deep and gradient boosting models \cite{ke2017lightgbm,gorishniy2021revisiting,voskou2024transformers}. We extracted several features related to each point, such as historical values for all parameters, historical values for neighboring areas, and location data. This model appeared to perform better; however, similar to previous attempts, we observed overfitting. While the model showed promising results for the earlier years of our validation splits, its performance deteriorated for more recent periods and test data.

Based on our preliminary analysis, we have drawn the following key insights:
i. The dynamics have significantly changed over the last ten years (Figure \ref{fig:side_by_side}) of the training data; thus, the older portion of the data has limited value.
ii. Most approaches attempted exhibited a tendency to overfit rapidly.
iii. When approaching the data globally, the rows-to-columns ratio is very low, making it difficult for deep models to achieve optimal results.
iv. Treating the task as a tabular data problem appears to be the most promising approach.

\subsubsection{The method}

In the proposed methodology, we maintained the tabular approach, treating each part of the map as a different row. However, we made two key changes: i) We used only the most recent period, completely dropping the majority of the provided data.ii) We employed a simpler model to avoid overfitting. More specifically we used the  Bayesian Ridge model.  The Bayesian Ridge model, at its core, is a simple linear model; however, it is further equipped to handle overfitting and small datasets more effectively. The vanilla ridge model is essentially a linear model that modifies the objective function by introducing an L2 regularization term, resulting in the objective function:
\[
\text{Objective} = \| Xw - y \|_2^2 + \lambda \| w \|_2^2
\]

The Bayesian Ridge model extends the vanilla ridge regression by incorporating Bayesian inference, which allows for the estimation of the posterior distribution of the model parameters. This approach enables the model to automatically determine the complexity of the model, balancing the fit to the data and the regularization term. Instead of using a fixed regularization parameter \(\lambda\), the Bayesian Ridge model treats \(\lambda\) as a random variable, deriving its value from the data itself. This results in a more flexible and robust model, especially useful for datasets with limited observations or potential overfitting issues.  For our model we used the default Scikit-Learn version \cite{scikit-learn}.

As input features, we utilized the historical Sea Surface Temperature Anomaly (SSTA) values from the past 12 months for specific locations, along with additional signals from the 8 directly neighboring locations (longitude and latitude \(\pm 1\)). For neighboring locations, we used the mean, maximum, and minimum values (per each month) instead of raw values, as we believe that, at least in the first approximation, the exact orientation does not significantly impact the results, and the land neighboring cases may affect the outcomes. Consequently, each input row comprises a total of \(12 + 3 \times 12 = 48\) features.

Our primary methodology was based on combining two models: one trained on the last 9 years of available input data, and the other on the last 3 years.

In addition to the primary model, we introduced two correction terms: a location- and season-aware term, \(C_{local}\), and a global fixed constant, \(C_{global}\). Our main model involved only one year of observation of local anomalies; hence, it did not consider the geographic location or the season. Although this information is crucial and different behavior is expected in tropical regions versus the Mediterranean or northern seas, and anomalies may also vary between summer and winter, we added the aforementioned \(C_{local}\) correction. This correction adjusted the output value using the mean SSTA (over the last 10 years of the training data) for the specific location and season. It is important to note that while location information is directly provided, the season is not. However, given that we have a global picture of sea temperature, inferring the corresponding season is trivial. While simply observing the average temperature in the northern versus southern hemispheres suffices to extract this, we aimed for more accurate predictions. Therefore, we assigned "seasonal pseudolabels" to the input data, artificially labeling it periodically every 12 rows. Using the global SST data normalized per row, we trained a simple 2-layer MLP and used it to match the test data to the corresponding seasonal information. Regarding the second correction term, we used a simple constant \((C_{global} = 0.1)\), which was found experimentally to improve test scores and is reasonable given the well-known global warming issue and the preliminary analysis \ref{fig:side_by_side} discussed earlier.

As a result, our overall model \footnote{code: https://github.com/avoskou/Diving-Deep} reads as:
\begin{equation}
f(x, \text{lt}, \text{lg}, s) = \frac{f_{\text{br}}(x|D_{\text{3y}}) + f_{\text{br}}(x|D_{\text{9y}})}{2} + \frac{C_{\text{local}}(\text{lt}, \text{lg}, s) + C_{\text{global}}}{2}
\end{equation}
where \(x \in \mathrm{R}^{48}\)  the corresponding row of the tabular input described above, \(s \in [1, 12]\) is the estimations of the season and  $ \text{lt},\text{lg}$ latitude, and  longitude.

\subsubsection{Post Competition Evaluation}

\subsubsection{Model Performance}

\begin{figure}
    \centering
    \includegraphics[scale=0.45]{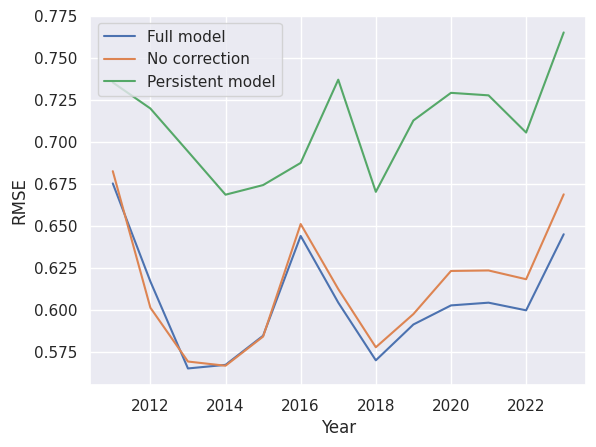}
    \label{all}
    \caption{Comparison of the proposed model with and without correction terms with the persistent model for 2010 to 2023}
\end{figure}

In Figure 2, we present the RMSE per year for the proposed model compared to the baseline (the persistent model). We include both the pre-correction prediction and the full proposed model. As shown, our approach is superior to the baseline for every single year, with the corrections adding measurable value, particularly in the later years. Crucially, in phase two, we managed to demonstrate an overall improvement of 0.1111 over the baseline.

{\bf Baltic Sea.} For the Baltic Sea case, we used Bayesian Ridge with a similar row structure, but this time using values from the last three Septembers rather than the previous 12 months, and without correction terms. As before, we combined two models trained on two different periods: the last 15 years and the last 10 years. The results are presented in the following table:

\begin{center}
\begin{tabular}{lll}
\toprule
Prediction & Latitude & Longitude \\
\midrule
0.592666 & 56 & 19.6875 \\
0.648081 & 58 & 19.6875 \\
0.425684 & 62 & 19.6875 \\
\bottomrule
\end{tabular}
\label{baltic}
\end{center}

\subsubsection{Conclusions and Future Work}

Using a relatively simple approach, we managed to surpass the baseline clearly and consistently; additional corrections provided a small yet measurable improvement.

The central observation was that the behavior and interaction of SSTA appear to change over time. Given this, an adaptive model that actively adapts to new data appears to be a promising direction for future work.

\subsection{Solution by Team UPB-DICE }

\subsubsection{Overview of the Approach.}
In the Diving Deep challenge, the aim was to investigate the predictability of global sea surface temperatures (SSTs) and sea surface temperature anomalies (SSTAs). Before designing suitable machine learning approaches to solve the problem, we first analyzed the data. In particular, we checked for missing values, and plotted the target distribution for different periods of time. With this, we realized that regression models are well suited for the task. We hence enumerated a number of candidate models, including LSTM~\cite{hochreiter1997long}, GRU~\cite{cho2014learning}, LightGBM~\cite{ke2017lightgbm}, XGBoost~\cite{chen2016xgboost}, RandomForest~\cite{liaw2002classification}, and CatBoost~\cite{prokhorenkova2018catboost}, which we then tested individually.
We provide more details in the next section.

\subsubsection{Main Approach.}
\label{main}
In the beginning, we opted for flattening all input features at all locations for every block of 12 months, and predicting the target values at the 5774 locations at once, but ran into RAM issues. Therefore, we changed the approach and applied a transposition operation after concatenating input features row-wise, then transposed the target values accordingly. With the four input features, each with 12 entries (12 months), and coordinates with 2 entries, our new training and validation data had an input space of 50 dimensions and a one-dimensional target. With this, the goal was to predict the target value at one location at a time. The data we obtained at this step had $4,832,838$ rows.

\subsubsection{Secondary Approach.}
\label{second}
In this approach, we add more features for training, validation and testing. This is achieved by considering the monthly average ($avg$) and standard deviation ($std$) of the target values (i.e., SSTA) for every location. Hence, for every month (i.e., January to December), we compute $avg$ and $std$ of SSTA values for 5774 locations throughout the years 1940 to 2010; this yields two matrices, each of shape $12\times 5774$. These matrices are transposed and concatenated to the initial feature matrices as described in the main approach above, resulting in a total of 74 input features.
\subsubsection{Models, Training, Validation, and Testing.}
\label{training}
With the data constructed above, we trained LSTM (2 recurrent layers followed by 5 linear layers), GRU (2 recurrent layers followed by 5 linear layers), LightGBM, and CatBoost in a 5-fold cross-validation setting. The other models were left out because their addition led to a decrease in predictive performance.
All models were ensembled for the final predictions: 0.42 for LightGBM, 0.42 for CatBoost, 0.06 for GRU, and 0.10 for LSTM. However, we noticed that by altering the weights slightly so that their sum is larger than $1$ and reducing the weight of GRU, we could further improve our score on the public leaderboard. Hence, we set our final weights as: 0.5 for LightGBM, 0.5 for CatBoost, 0.12 for LSTM, and 0.07 for GRU. Our code to reproduce the results can be found at \url{https://github.com/arnabsharma91/deepSSTA}.
We expect that further adjustments of the weights would yield even better performance.

\subsubsection{Main Results.}
\begin{figure}
\centering
    \subfigure{\includegraphics[scale=0.403]{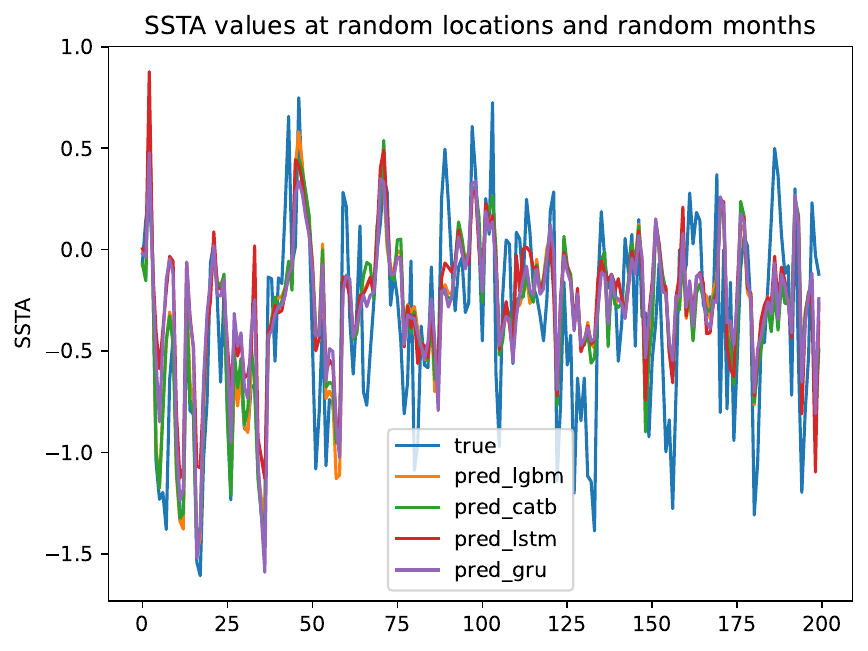}}
    \subfigure{\includegraphics[scale=0.403]{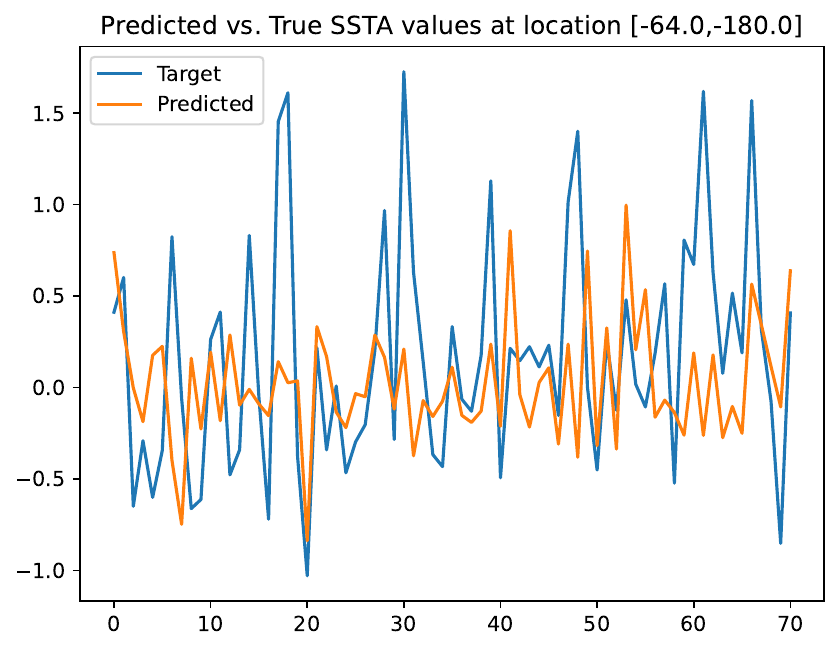}}
\caption{{\bf Left:} Visualization of model performance on the validation dataset, {\bf Right:} Predicted vs. target in testing phase}
\label{fig:validation}

\end{figure}

Figure~\ref{fig:validation} {\bf(Left)} shows the performance of each model on the 4th fold of the 5-fold splits. Note that the performances on the other folds follow a similar trend. Since the validation sets were obtained via a random split, we do not know to which month or location the predictions correspond. However, the figure suggests that our predictions follow the direction of the true target in most cases.

In Figure~\ref{fig:validation} {\bf (Right)}, we compare our predictions to the true target during the testing phase at location [-64.0, -180.0]. It is hard to predict the true target as it frequently alternates between low and high values. This is why our predictions tend to be close to 0 so the overall error can be minimized. Nonetheless, our predictions follow the overall trends, and we can predict values close to the target for specific months, see for example month $20$. This suggests that our models can serve as strong predictors or help to validate predictions made by other approaches.

\subsubsection{Predictions for September 2024.}
We are given input data till December 2023 but our models were trained to predict for 3 months in advance. We hence have to proceed sequentially: the given data can be used to predict SSTA values for January, February, and March 2024. Then, we use the newly predicted values to predict for April, May, and June 2024. Finally, we use input data for [July 2023, ..., June 2024] to predict for September 2024. Note that our models can predict only SSTA values. Therefore, only SSTA values change in the input to our models, while values for the other variables remain unchanged. The SSTA values predicted for September 2024 are reported in the table below. Please refer to the notebook \verb|Predict_September2024.ipynb| for implementation details.
\begin{center}
\begin{tabular}{lll}
\toprule
Prediction & Latitude & Longitude \\
\midrule
0.716 & 56 & 19.6875 \\
0.460 & 58 & 19.6875 \\
-0.216 & 62 & 19.6875 \\
\bottomrule
\end{tabular}
\end{center}

\section{Outlook}

The outcomes of the Diving Deep Challenge reveal advancements in the predictability of SSTAs through a diverse range of methodological approaches. The top-performing teams, Team randomguy and Team UPB-DICE, showcased techniques that leveraged both conventional machine learning and deep learning approaches, which highlight the importance of model simplicity, data selection, and ensemble techniques in achieving high predictive accuracy, particularly in the context of challenging SSTA prediction.

Looking forward, there are several key areas where future challenges could be directed. One promising direction is the exploration of multi-model ensembles that combine predictions from different models to improve overall forecast accuracy and robustness. Another avenue is expanding the scope of SSTA prediction to include other relevant oceanographic and atmospheric variables, such as ocean currents or wind patterns, which could provide a more comprehensive understanding of the factors driving SSTA changes. SSTAs at different depths, along with data from oceanic buoys, could be particularly valuable when combined with recent advances in deep learning-based meteorological models, especially for long-term SSTAs prediction \cite{lam2023graphcast}. Furthermore, the challenge demonstrated the need for enhanced data integration strategies, particularly when working with limited or sparse datasets. Techniques such as data augmentation and transfer learning could play a positive role in improving model performance. Future challenges may also benefit from incorporating real-time data streams and/or high-quality local data, allowing models to adapt to the most current and/or local information and providing relevant forecasts.

The insights gained from this challenge pave the way for future endeavors to explore the data-driven predictability of SSTAs and enhance our understanding of the Earth's climate system. By continuing to push the boundaries of what is possible in climate forecasting, we can better prepare for the challenges posed by a changing climate and work towards more sustainable and resilient ocean ecosystems.

\begin{credits}
\subsubsection{\ackname} We would like to acknowledge that this competition is supported by the TAIAO project CONT-64517-SSIFDS-UOW (Time-Evolving Data Science / Artificial Intelligence for Advanced Open Environmental Science) funded by the New Zealand Ministry of Business, Innovation, and Employment (MBIE), URL: \href{https://taiao.ai/}{https://taiao.ai/}.
\end{credits}
%
%
%
\bibliographystyle{splncs04}
\bibliography{paper}
%
\end{document}